\documentclass[runningheads]{llncs}
\usepackage{graphicx}
\usepackage{booktabs}
\usepackage{amsmath}
\usepackage{amsfonts}
\begin{document}
\title{AU-vMAE: Knowledge-Guide Action Units Detection 
via Video Masked Autoencoder}
\titlerunning{AU-vMAE}
\author{Qiaoqiao Jin\inst{1}\orcidID{0009-0006-7733-4684} \and
Rui Shi\inst{1}\orcidID{0009-0008-7335-7218} \and
Yishun Dou\inst{2}\orcidID{0009-0008-8345-8258} \and
Bingbing Ni\inst{1}\orcidID{0000-0001-5582-577X}\thanks{Corresponding author: Bingbing Ni.} }
\authorrunning{Q. Jin et al.}
%
\institute{Shanghai Jiao Tong University, Shanghai 200240, China \\ \email{\{jinqiaoqiao, nibingbing\}@sjtu.edu.cn}
\and
Huawei 
}

\maketitle              
\begin{abstract}
 Current Facial Action Unit (FAU) detection methods generally encounter difficulties due to the scarcity of labeled video training data and the limited number of training face IDs, which renders the trained feature extractor insufficient coverage for modeling the large diversity of inter-person facial structures and movements. To explicitly address the above challenges, we propose a novel video-level pre-training scheme by fully exploring the multi-label property of FAUs in the video as well as the temporal label consistency. At the heart of our design is a pre-trained video feature extractor based on the video-masked autoencoder together with a fine-tuning network that jointly completes the multi-level video FAUs analysis tasks, \emph{i.e.} integrating both video-level and frame-level FAU detections, thus dramatically expanding the supervision set from sparse FAUs annotations to ALL video frames including masked ones. Moreover, we utilize inter-frame and intra-frame AU pair state matrices as prior knowledge to guide network training instead of traditional Graph Neural Networks, for better temporal supervision. Our approach demonstrates substantial enhancement in performance compared to the existing state-of-the-art methods used in BP4D and DISFA FAUs datasets. All source code will be released once upon acceptance.
\end{abstract}
\section{Introduction}
The Facial Action Coding System (FACS)~\cite{FACS}, developed by Friesen and Ekman in 1978, uses Facial Action Units (FAUs) to represent human facial expressions based on the movements of facial muscles. 
There are a total of 46 FAUs that can be seen in the human face, each of which corresponds to a specific muscle or group of muscles, combined to control expressions. 
Compared to emotion-based categorical models, FAUs provide a more comprehensive and objective way of describing facial expressions~\cite{survey}.
Previously, the detection of FAUs was primarily limited to image-level analysis utilizing convolutional neural networks (CNNs)~\cite{DRML,JAA,EAC-NET} or graph neural networks (GNNs)~\cite{ME-graph,CAF-NET} for feature extraction. Solely exploring features at the image level has difficulties achieving higher classification accuracy given the limited volumes of annotated Action Units~(AU) data. Since videos can capture the dynamic temporal evolution of facial muscle activations underlying various expressions, the incorporation of video modality provides the potential for enhanced FAU status modeling and more robust facial expression recognition. However, a salient challenge stems from the limited availability of temporally annotated video samples within existing FAU datasets. Notably, widespread datasets such as BP4D~\cite{BP4D} and DISFA~\cite{DISFA} contain merely hundreds of videos, rendering the volumes insufficient for robust network training towards precise identification and detection of the diverse spectrum of FAUs.


\begin{figure}[t]
    \centering
  \includegraphics[width=0.6\linewidth]{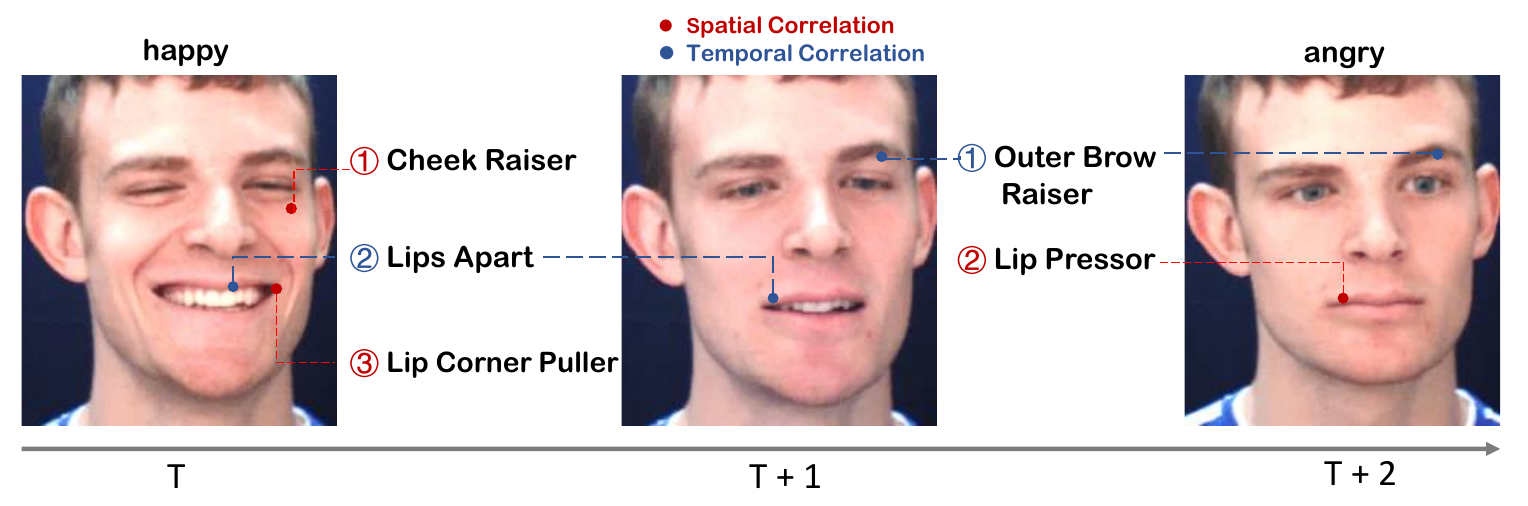}
  \vspace{-0.3cm}
  \caption{AU labels exhibit spatial and temporal correlations due to expressions activating FAUs, inducing spatial relations, and changing musculature associating temporally distinct transitions.}
  \label{fig:intro}
  \vspace{-0.5cm}
  
\end{figure}

We use video data for FAU detection because the human face muscles exhibit frequent and robust correlations over time. For instance, when a person smiles, the muscles around the cheek (AU6) and mouth (AU12) move synchronously. Typically, these two muscles are in a relaxed state (\emph{i.e.} no smile). When activated, AU12 pulls the corners of the mouth upwards, resulting in a slight smile. Both muscles then become activated, leading to a full smile.
This clear sequential movement of the cheek and mouth muscles during a smile demonstrates the spatial and temporal relationships between action units. The insight from such relationships motivates us to leverage rich spatiotemporal correlation information to expand self-supervised learning during detector training, aiming for more robust and accurate feature modeling and recognition.


To maximally explore the spatiotemporal structural information from video data in the FAU detection task, we propose a novel training scheme that successfully mines dense self-supervision signals from the spatiotemporal dependencies from sparsely FAU-labeled video data, called \emph{AU-vMAE} method, constructed within the popular video masked auto-encoder (videoMAE)~\cite{videoMAE} training framework.
This approach comprises two main components. Firstly, a vision transformer~\cite{ViT} is utilized as an encoder to reconstruct the masked face videos. Thus we can efficiently recover the underlying temporal and spatial structures within the video, thereby facilitating the extraction of meaningful features from the data. Then we utilize the pre-trained encoder to detect each frame's FAUs label via a multi-label binary classification task. Considering the temporal and spatial correlation between two FAUs, we employ finite state machine (FSM) models to evaluate the intra-frame (spatial) and inter-frame (temporal) relations of AU pairs. The intra-frame state machine stands for the probability of co-occurrence for each pair of AUs to constrain their spatial occurrence. The inter-frame state machine calculates the probability of state transition for each pair of AU labels, thereby constraining the temporal occurrence of AU pairs. Moreover, our AU-vMAE introduces a multi-level training scheme for enhancing FAU detection, \emph{i.e.} by performing both AU label prediction on the video frame by frame (video-level), as well as on the entire video using extracted frames (frame-level). Additionally, it can detect FAUs on randomly masked videos frame-by-frame (patch-level).  The experimental results show a 3.2\% enhancement in F1-score on the BP4D dataset and a 5.8\% enhancement on the DISFA dataset compared to the state-of-art method.

\begin{figure*}[t]
    \centering
  \includegraphics[width=1.0\linewidth]{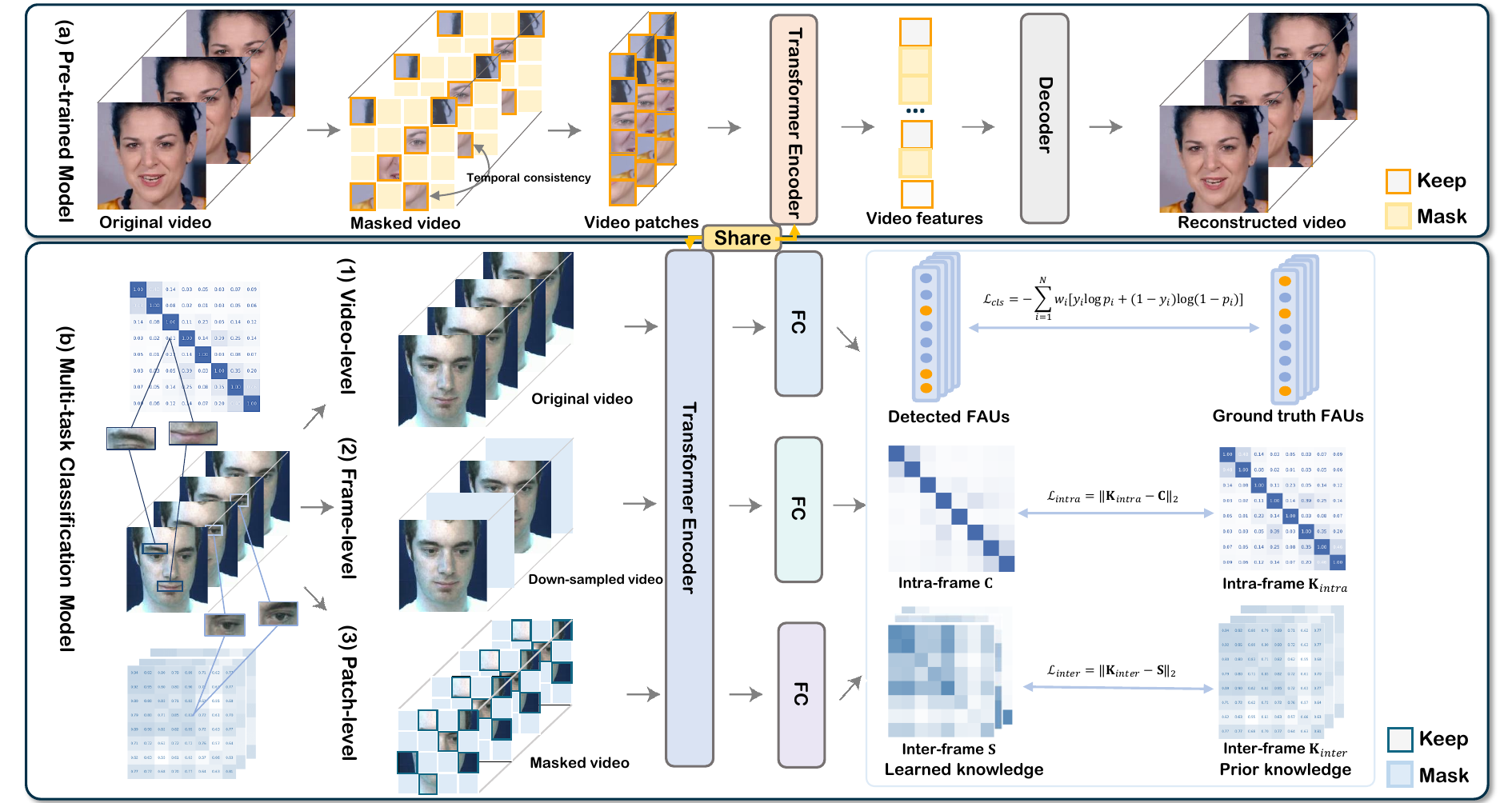}
  \caption{The AU-vMAE framework comprises two primary modules: (a) \textit{ A pre-trained model} is developed by reconstructing facial videos with masks to extract video features. (b) This pre-trained encoder is then applied to \textit{three downstream subtasks}: 1) \textit{video-level FAU detection} which processes all frames of a video to predict AU labels for each frame; 2) \textit{frame-level FAU detection} using equidistantly sampled frames to predict AU labels for the entire video, and 3) \textit{patch-level FAU detection} where randomly masked frames are used to predict AUs frame-by-frame.}
  \vspace{-0.5cm}
  \label{fig:overview}
\end{figure*}

\section{Related Work}
\subsection{Facial Action Units Detection}
Early FAU detection approaches rely on landmark identification to determine regions of interest and train classifiers using either neural networks or Support Vector Machines (SVM)~\cite{SVM}. To enhance detection accuracy, attention maps have been introduced to aid CNN in locating AU labels, such as EAC-Net~\cite{EAC-NET} and JAA-Net~\cite{JAA}. Recent researchers have introduced novel methodologies through which graph neural networks are employed to refine features and generate structures. For instance, Zhang et al.~\cite{heatmap} applied a heatmap regression-based approach, while Song et al.~\cite{Hybrid} proposed a performance-driven hybrid message-passing neural network with optimized structures, abbreviated as HMP-PS. ME-GraphAU~\cite{ME-graph} differentiates itself from previous GCN by training multi-dimensional edge features for every pair of AUs, resulting in improved performance when compared to previous techniques. Different from the previous studies aiming to learn ``pointwise feature" knowledge, we incorporate a state machine as prior knowledge which is more proper to enhance the precision of our knowledge description regarding the temporal-spatial status of the AU labels. 

\subsection{Video Feature Extraction}

Image level feature extraction has been widely used for AU detection in prior research~\cite{DBLP:conf/cvpr/JacobS21,ME-graph,Aff-wild2}, such as landmark detection, attention mapping, and aggregating node features across AU labels. Beyond exploring informative features, pre-trained networks~\cite{ME-graph,Face-MAE,yin2023multi} on large-scale datasets~\cite{deng2009imagenet} are leveraged to empower the models by extracting discriminative image representations.
In particular, the Mask AutoEncoder (MAE) framework \cite{ImageMAE} has advanced the accuracy of FAU detection \cite{Face-MAE,DBLP:journals/corr/abs-2303-10644}, which is primarily designed for static images and focuses less on capturing the temporal relationships between AUs. 

To effectively leverage temporal dependencies in limited data, current approaches~\cite{DBLP:conf/cvpr/0002TJW21,videoMAE} often employ a two-stage pre-training and fine-tuning scheme. Among them, VideoMAE~\cite{videoMAE} has demonstrated outstanding capability for the temporal modeling of videos. This novel framework facilitates video reconstruction through a downsampling masking process that extracts spatiotemporal features. 
Its adept handling of temporal and spatial dynamics, coupled with effective feature extraction ability, gives it the potential for high classification accuracy in FAU detection tasks, even with limited data.

\section{Method}
\subsection{Overview}
To maximally explore the spatiotemporal structural information of FAUs, we follow a two-step approach to perform FAU detection. First, a base version of videoMAE~\cite{videoMAE} is pre-trained for video reconstruction on large-scale face datasets, ending up with a video encoder facilitating extracting video features.
Second, we present a fine-tuning network for FAU detection with multi-level inputs, integrating prior intra-frame and inter-frame knowledge to constrain the distribution of predicted AU labels. The fine-tuned network has a strong ability to capture temporal and spatial information of video data, allowing us to predict AU labels for every frame of a video, whether the input is individual frames (video-level), downsampled frames (frame-level), or masked frames (patch-level).

\subsection{Pre-trained Feature Extractor}
\label{sec:pretrain}
Different from prior research on image-level FAU detection~\cite{JAA,EAC-NET,DBLP:conf/cvpr/JacobS21}, our study concentrates on video-level FAU detection, where the network processes sequences of temporally adjacent frames. To enhance feature extraction with limited AU-labeled data, we employ large-scale unlabeled video datasets for pre-training using the VideoMAE~\cite{videoMAE} architecture, which performs self-supervised masked video reconstruction. As depicted 
in Fig.~\ref{fig:overview}, the facial video frames are subjected to temporal and spatial compression via downsampling and mask. Subsequently, a subset of tokens undergoes tube masking with a high ratio (90\%), while the remaining tokens are fed into an encoder based on Vision Transformer. By decoding these visible tokens, we can reconstruct the video via L2 loss: 
\begin{small}
\begin{equation}
    \mathcal{L} = \frac{1}{N}\sum_{i = 1}^N \left[\frac{1}{\omega_i}\sum_{t\in \omega_i}||I_i(t)-\hat{I}_i(t)||_2 \right],
\end{equation}
\end{small}
where $t$ is the token index, $\omega_i$ is the set of masked tokens of the $i$th frame, $I_i$ is the $i$th input frame, and $\hat{I}_i$ is the $i$th reconstructed frame. 
To align the frame downsampling rate across pre-training and fine-tuning phases, we adjust the pre-trained model's downsampling rate to suit various fine-tuning tasks.
 For video-level tasks and patch-level tasks, no downsampling is applied. For frame-level tasks, the downsampling rate is set to 4. This allows the pre-trained model to better retain its feature extraction ability learned during the fine-tuning phase.

Ultimately, we obtain a pre-trained model that adequately extracts temporal and spatial video features, which benefits from the large facial dataset, thus avoiding overfitting issues when relying solely on the available FAUs dataset in the fine-tuning phase. Two main operations improved the model's ability to extract temporal and spatial features: \textbf{\textit{\romannumeral1)}} \textit{Tube masking} enables it to proficiently determine the masked patch information within frames; and \textbf{\textit{\romannumeral2)}} \textit{Temporal downsampling} allows for the establishment of robust semantic associations between different frames, resulting in a superior representation of the video contents.

\subsection{Knowledge-Guided AU Classification}
\subsubsection{Multi-level AU Classification. }
The pre-trained model in Sec.~\ref{sec:pretrain} is the basis for our downstream task, FAU detection. We frame it as a multi-label binary classification task and further divide it into three subtasks according to the input level: \textbf{\textit{\romannumeral1)}} \textbf{video-level FAU detection}, taking all frames of a given video as input; \textbf{\textit{\romannumeral2)}} \textbf{frame-level FAU detection}, taking downsampled frames as input; and \textbf{\textit{\romannumeral3)}} \textbf{patch-level FAU detection}, taking tube-masked all frames as input. Regardless of the input level, the network always predicts AU labels for every frame of the video.

In subtask 1, FAU detection is performed frame-by-frame, similar to image FAU detection. Therefore, we compare our method with previous image-level approaches to highlight the efficiency of AU-vMAE. Subtask 2 only requires 1/4 of the total number of video frames as input to predict AU labels for all frames, which significantly improves computation efficiency for FAU detection. The redundant timing information between adjacent frames and the time-invariant nature of AU changes contribute to the network efficiency. In subtask 3, 50\% video patches are randomly blocked out to simulate real-life scenarios where the face is partially obscured. The blocked areas can be implicitly completed in features by the pre-trained encoder, thereby achieving precise AU prediction.

To address the issue of the imbalanced label distribution in AU datasets~\cite{BP4D,DISFA}, we calculate weights for each label using occurrence rates in the training set, defined as $w_i = N(1/r_i)/\sum^N_{j=1}(1/r_j)$, where $r_i$ is the occurrence rate of \textit{i}th AU. These weights ($w_i$) are incorporated into the loss for the classification tasks:
\begin{equation}
\label{equ:L_cls}
    \mathcal{L}_{cls}= -\sum^N_{i=1}w_i\left[y_i \log p_i + (1-y_i) \log (1-p_i)\right],
\end{equation}
where $N$ represents the number of AUs, $p_i$ signifies the predicted probability which is sigmoid activations on the model outputs, and $y_i$ indicates the binary ground truth of the $\text{AU}_i$.
\begin{figure}[t]
  \centering
  \includegraphics[width=0.5\linewidth]{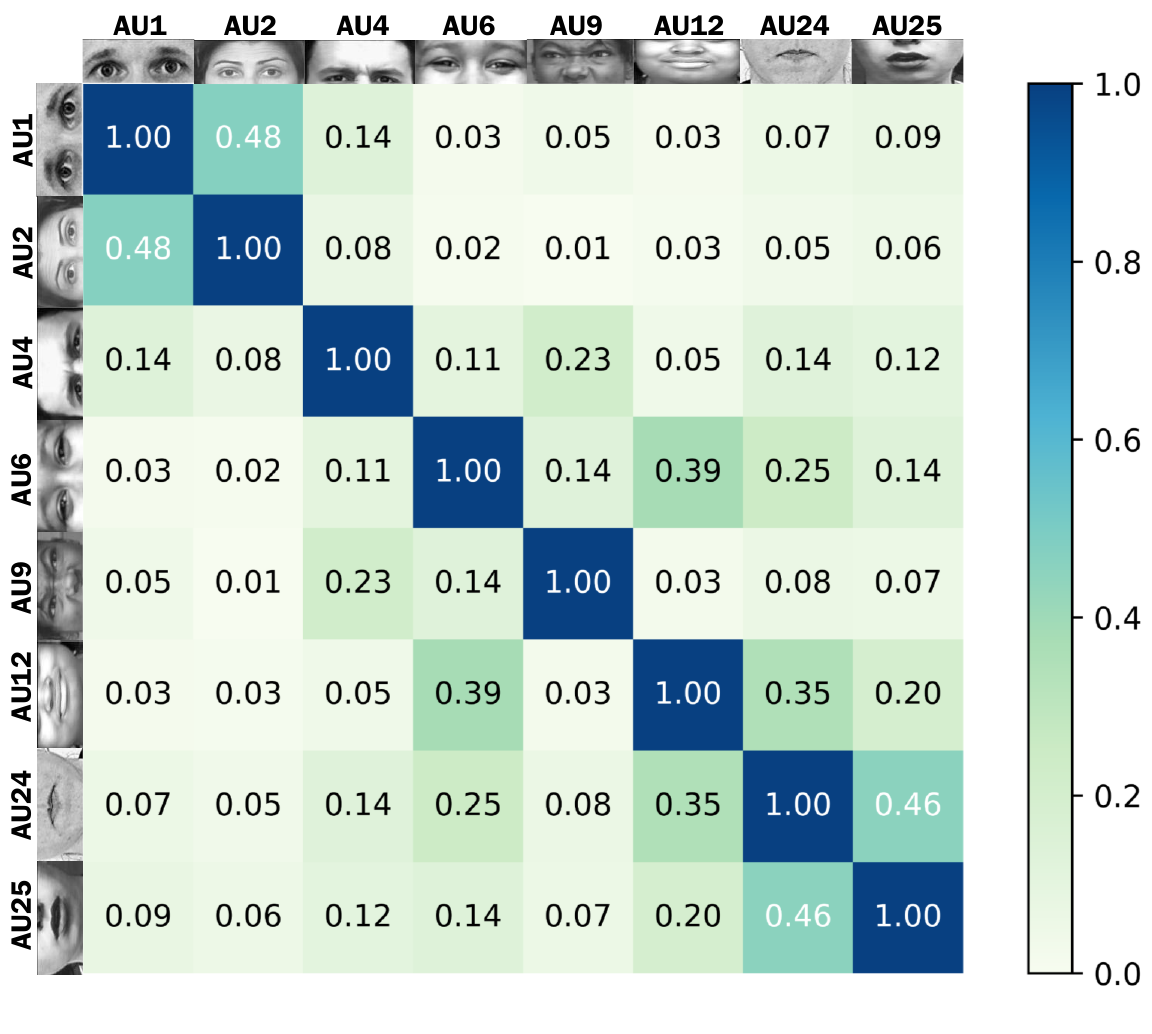}
  \vspace{-0.3cm}
  \caption{Co-occurrence matrix of DISFA~\cite{DISFA} dataset. The co-occurrence matrix displays the likelihood of two labels appearing together, with each element corresponding to a pair of AU labels.}
  \label{fig:Co-occurrence}
  \vspace{-0.3cm}

\end{figure}

\subsection{AU-pair Finite State Machine}
Prior researches~\cite{ME-graph,DBLP:conf/cvpr/JacobS21,Hybrid} have recognized the interdependence among AUs. However, these approaches entail calculating the state space comprised of all AU labels and subsequently utilizing methods such as KNN for state selection, which is redundant and complicated. To achieve a more precise characterization of AU labels, we employ \textit{Finite State Machines} (FSM) to describe the state of each AU pair. Specifically, our approach combines every two AU labels as their state. The decision is based on the fact that the state space engendered by a dozen AU labels is vast and redundant, while the state space formed by the AU pair is more compact and rational. Moreover, knowledge represented by two AU labels tends to be statistically regular, whereas states composed of multiple labels become over-specific and lose their expressiveness.

\subsubsection{Intra-frame AU-pair Finite State Machine.}
In previous works~\cite{DRML,JAA}, correlation matrices are effective in extracting intra-frame features from facial regions where AUs tend to co-occur. Here, we define the intra-frame knowledge $\mathbf{K}_{intra} \in \mathbb{R}^{N \times N}$, where $N$ is the number of AU labels, as the state co-occurrence matrix of AU pairs in the training dataset. To get $\mathbf{K}_{intra}$, we encode the intra-frame states of each pair of AU labels (\emph{e.g.} AU1 and AU2) as binary values: \textbf{`00'} (indicating both labels are negative), \textbf{`01'} (indicating AU1 is negative and AU2 is positive), \textbf{`10'} (indicating AU1 is positive and AU2 is negative), and \textbf{`11'} (indicating both labels are positive). Each element in $\mathbf{K}_{intra}$ represents the probability of co-occurrence between a pair of AU labels. We define the co-occurrence probability as the likelihood that the AU labels state \textbf{`11'} (\emph{e.g.} $p(i=1,j=1)$) under the condition where at least one AU label is activated (state `10' or `01' or `11'). We remove the state `00' because it constitutes the majority of the labels, and including it would result in very small values in the elements of $\mathbf{K}_{intra}$, which would not effectively guide network training. Therefore the $\mathbf{K}_{intra}$ can be expressed as:
\begin{small}
 \begin{equation}
 \label{equ:equ3}
   \mathbf{K}_{intra} = \frac{p(i=1,j=1)}{p(i=1,j=0) + p(i=0,j=1)+ p(i=1,j=1)}.
\end{equation}
\end{small}

During training, predicted probabilities $\mathbf{p}\in\mathbb{R}^{b\times N}$ are obtained by applying sigmoid activations to the model outputs, where $b$ represents the batch size and $N$ represents the number of AU labels. To obtain the final predicted labels in the form of binary states (0 or 1) for the AU, we compare $\mathbf{p}$ with a threshold of 0.5. Then the learned intra-frame knowledge $\mathbf{C}$ can be expressed mathematically as: 
\begin{small}
\begin{equation}
\label{equ:equ4}
\mathbf{C} = \frac{\left[\mathbf{p}>0.5\right]^T \left[\mathbf{p}>0.5\right]}{b\mathbf{I}-(\mathbf{I}-\left[\mathbf{p}>0.5\right])^T(\mathbf{I}-\left[\mathbf{p}>0.5\right])}. 
\end{equation}
\end{small}
Here the Iverson bracket indicator function $\left[\mathbf{p}>0.5\right]$ evaluates to $1$ when $\left[\mathbf{p}>0.5\right]$ and $0$ otherwise. 
However, Eq.~\ref{equ:equ4} is non-differentiable due to the discrete operation of assigning binary states (0 or 1). As a solution, a gradient-crossing operation~\cite{TakikawaET0MJF22} is adopted to skip the gradient of this step to make the entire network operation differentiable, which can be expressed as $[\mathbf{p}>0.5] - \mathbf{p}.detach() + \mathbf{p}$.
 
In the training process, we evaluate the learned intra-frame knowledge $\mathbf{C}$ each batch while measuring its L2 distance from the $\mathbf{K}_{intra}$ as:
\begin{small}
\begin{equation}
        \mathcal{L}_{intra}= \Vert \mathbf{K}_{intra} - \mathbf{C}\Vert_2.
\end{equation}
\end{small}

Figure~\ref{fig:Co-occurrence} illustrates the resulting co-occurrence matrix based on our encoding scheme applied to the DISFA dataset~\cite{DISFA}. 
The high co-occurrence probability between AU1 (Inner Brow Raiser) and AU2 (Outer Brow Raiser) indicates their frequent simultaneous activation. Additionally, the co-activation probability of AU6 (Cheek Raiser) and AU12 (Lip Corner Puller) is 0.39, as observed from the statistical co-occurrence matrix. The observed high correlation between AU6 and AU12 is in line with the common sense of the muscles involved in cheek and eye movements during laughter.

\begin{figure}[t]
  \vspace{-0.2cm}
  \centering
  \includegraphics[width=0.85\linewidth]{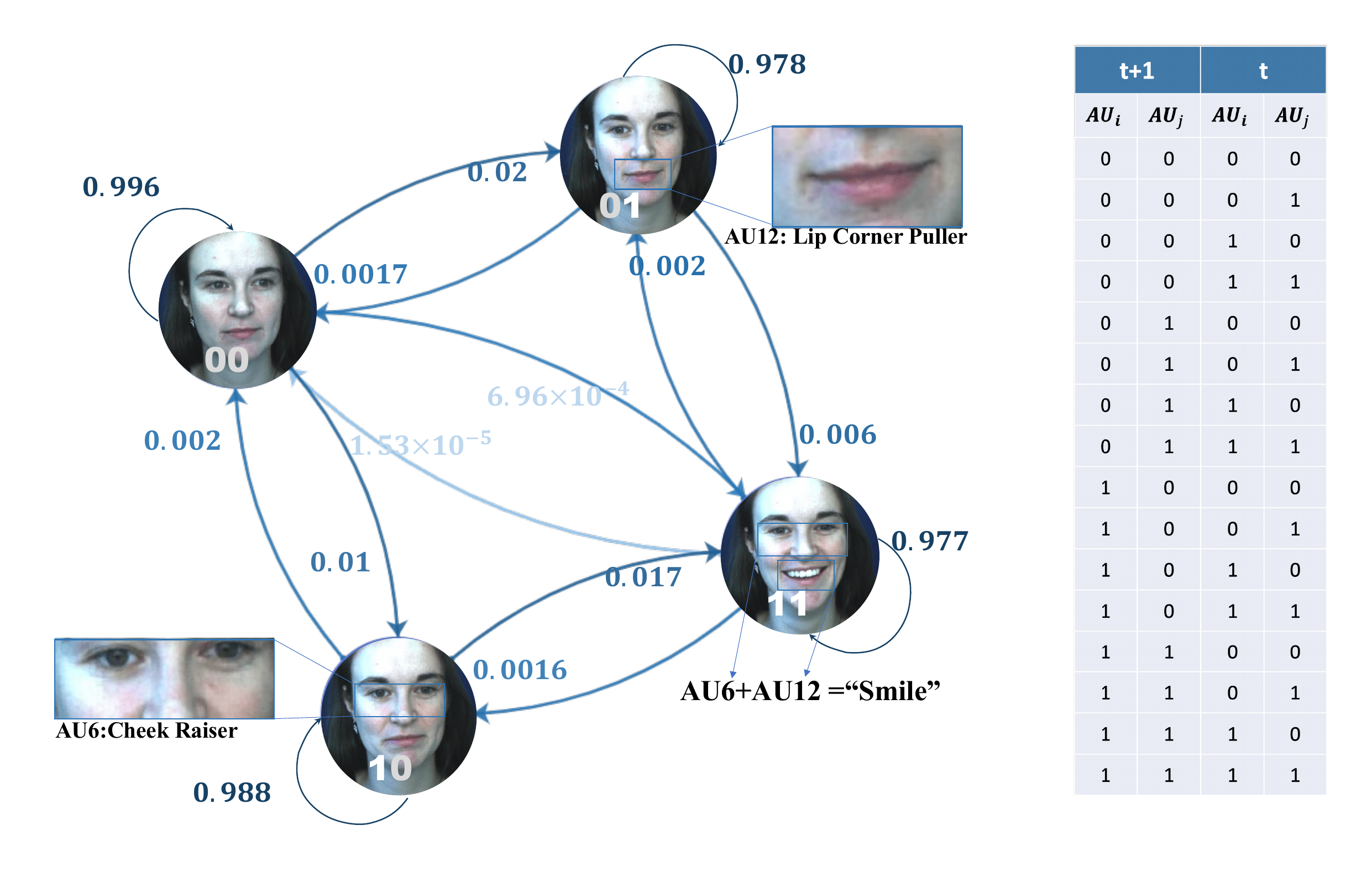}
  \vspace{-0.6cm}
  \caption{Finite state machine (FSM) of two AU labels. At any moment $t$, two AU label pairs ($i$ and $j$) can have four possible states~(00, 01, 10, and 11). The table presents 16 potential state transitions from time $t$ to time $t+1$. Their corresponding probabilities are visualized in the graph.}
  \label{fig:FSM}
  \vspace{-0.4cm}
\end{figure}

\subsubsection{Inter-frame AU-pair Finite State Machine.}
Facial expressions frequently exhibit a coherent and sequential pattern, such as the transition from an initial state of ``smiling" to a subsequent state of ``laughing".
We propose modeling the temporal transitions of AU pairs as 4-bit binary sequences in a finite state machine (FSM) framework, indicating the presence of a distinct mechanism governing these transitions.
As shown in Fig.~\ref{fig:FSM}, we observe four current states (00, 01, 10, and 11) for each pair of AU labels at any moment $t$ during temporal video analysis. These states can transition into one of four subsequent states at the next moment $t+1$. Consequently, the FSM exhibits 16 states to represent all possible transitions for each AU pair, as shown in the table of Fig.~\ref{fig:FSM}. To ensure controlled temporal state changes during training, we calculate the probabilities associated with the state transitions of all AU pairs. We define 
inter-frame knowledge $\mathbf{K}_{inter} \in \mathbb{R}^{N \times N \times 16}$ as a probability matrix of state transitions, where $N$ is the number of AU labels. Each element of the matrix, denoted by $(i, j)$, represents the probability of the 16 states corresponding to AU label $i$ and AU label $j$. The main objective is to ensure that the learned inter-frame knowledge matrix $\mathbf{S}$ approximates the established prior knowledge matrix $\mathbf{K}_{inter}$ accurately.
  
In practice, we tabulate the states of consecutive frames for each AU pair in the training set and estimate the probability of each state based on the statistics. This probability serves as our inter-frame knowledge. To learn the inter-frame knowledge in a differentiable manner, we propose a method where we convert the state index $s$ (ranging from 0 to 15) into its binary representation $s_3s_2s_1s_0$. We define the state function $\mathcal{D}$ as follows:
\begin{equation}
\mathcal{D}_{k}\left( x\right) = 
\begin{cases}
1-x,s_k=1\\
x,s_k=0
\end{cases}
\end{equation}
\begin{figure}[t]
  \centering
  \includegraphics[width=0.9\linewidth]{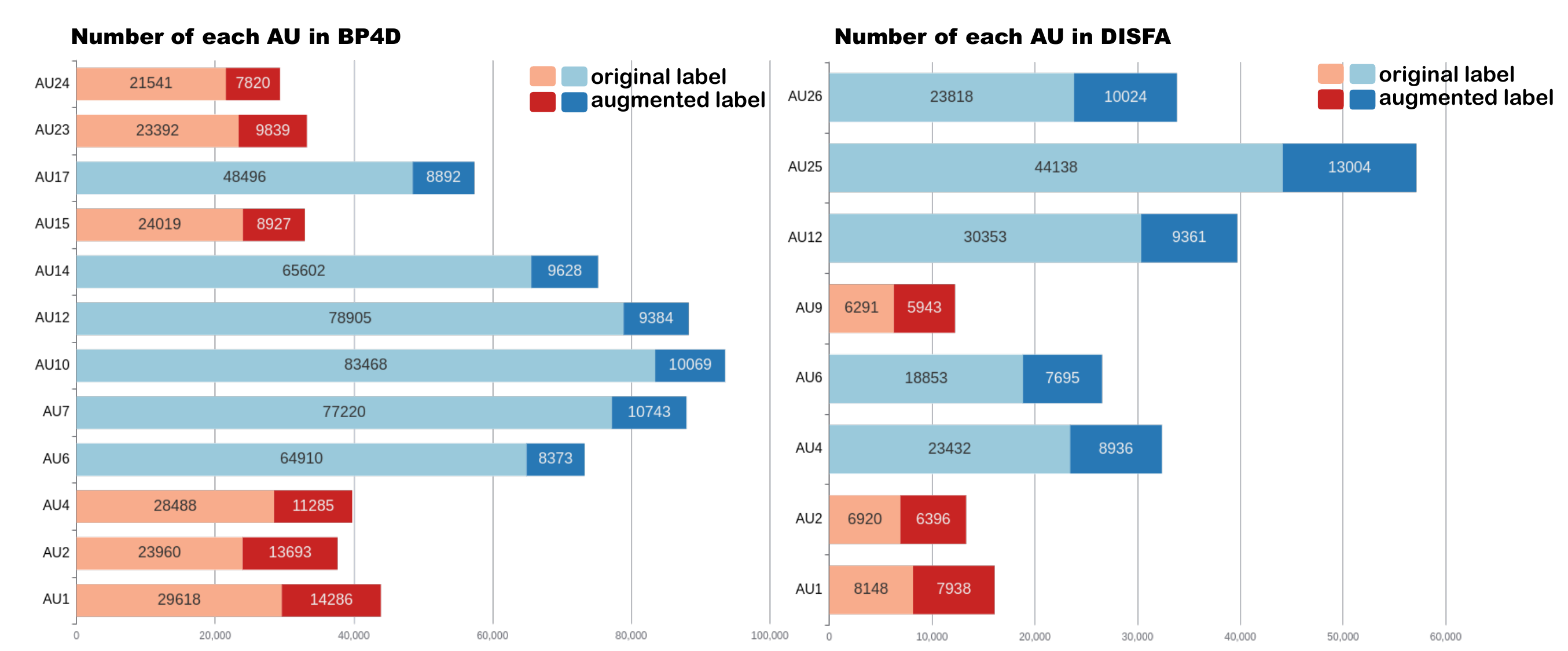}
  \caption{The distribution of the original AU label distribution and the augmented AU label distribution of BP4D and DISFA datasets.}
  \label{fig:chart}
  \vspace{-0.3cm}
\end{figure}
By utilizing the state function, we can derive the state matrix using AU labels in $t$th and $t+1$th frame.
The element-wise expression of the state matrix $\mathbf{S}$ is
\begin{equation}
    \mathbf{S}[i, j, s] = \mathcal{D}_{0}(p_i^t)\mathcal{D}_{1}(p_j^t)\mathcal{D}_{2}(p_i^{t+1})\mathcal{D}_{3}(p_j^{t+1}).
\end{equation}
The final loss of the inter-frame state machine is:
\begin{equation}
        \mathcal{L}_{inter}= \Vert\mathbf{K}_{inter} - \mathbf{S}\Vert_2.
\end{equation}
Finally, the training loss of our neural network is:
\begin{equation}
    \mathcal{L} =  \lambda_{cls} \mathcal{L}_{cls} +  \lambda_{intra}\mathcal{L}_{intra} + \lambda_{inter}\mathcal{L}_{inter},
\end{equation}
where $\lambda_{cls}$, $\lambda_{intra}$, and $\lambda_{inter}$ balance the weight of three loss terms. We experimentally set $ \lambda_{cls}$ = 1.0, $\lambda_{intra}$ = 0.01, and $\lambda_{inter}$ = 0.01.

\section{Experiment}
\subsection{Experimental Setup}
 
\subsubsection{Datasets.}
During pre-training process, 6 video facial datasets without AU labels are utilized, which are VoxCeleb2~\cite{Vox}, CelebV-HQ~\cite{celebvhq}, FaceForensics~\cite{roessler2018faceforensics}, VFHQ~\cite{xie2022vfhq}, and MEAD~\cite{mead}. 
We jointly train these five datasets to enhance the generalizability of our pre-trained model, with a total of about 0.5 million videos available. All these videos have undergone face detection and contain only facial content. Before being input into the network, they are resized to $224\times224$.

\begin{table*}[tb]
\centering
\small
\caption{F1 scores, expressed in percentages (\%), are measured for 12 AUs on the BP4D dataset.The best and second best results of each column are indicated with bold font and underlined respectively.}
\resizebox{.95\textwidth}{!}{%
\begin{tabular}{l|cccccccccccc|c}
\toprule
\multicolumn{1}{l|}{\textbf{Method}} &
  \multicolumn{12}{c|}{\textbf{AU} }&
  \multicolumn{1}{c}{\textbf{Avg.}} \\ 
\multicolumn{1}{l|}{} &
  \multicolumn{1}{c}{1} &
  \multicolumn{1}{c}{2} &
  \multicolumn{1}{c}{4} &
  \multicolumn{1}{c}{6} &
  \multicolumn{1}{c}{7} &
  \multicolumn{1}{c}{10} &
  \multicolumn{1}{c}{12} &
  \multicolumn{1}{c}{14} &
  \multicolumn{1}{c}{15} &
  \multicolumn{1}{c}{17} &
  \multicolumn{1}{c}{23} &
  \multicolumn{1}{c}{24} &
  \multicolumn{1}{|c}{}\\
  \midrule
DRML~\cite{DRML} & 36.4 & 41.8 & 43.0 & 55.0 & 67.0 & 66.3 & 65.8 & 54.1 & 33.2 & 48.0 & 31.7 & 30.0 & 48.3 \\
EAC-Net~\cite{EAC-NET} & 39.0 & 35.2 & 48.6 & 76.1 & 72.9 & 81.9 & 86.2 & 58.8 & 37.5 & 59.1 & 35.9 & 35.8 & 55.9 \\
JAA-Net~\cite{JAA} & 47.2&44.0 &54.9 &77.5 & 74.6 & 84.0 & 86.9 & 61.9 & 43.6 & 60.3 & 42.7 & 41.9 & 60.0 \\

SEV-Net~\cite{SEV-net}& \underline{58.2}& \underline{50.4}& 58.3 &\bf 81.9& 73.9 & \bf 87.8 & \underline{87.5} & 61.6 &52.6 &62.2& 44.6 & 47.6 & 63.9 \\

FAUDT~\cite{DBLP:conf/cvpr/JacobS21}& 51.7& 49.3& \bf 61.0 &77.8& 79.5 & 82.9 & 86.3 & 67.6 & 51.9  &63.0& 43.7 & \underline{56.3} & 64.2 \\

HMP-PS~\cite{Hybrid}& 53.1& 46.1& 56.0 &76.5& 76.9 & 82.1 & 86.4 & 64.8 & 51.5  &63.0& \underline{49.9} & 54.5 & 63.4\\

ME-AU~\cite{ME-graph} &52.7& 44.3& \underline{60.9} &\underline{79.9}& \underline{80.1}& \underline{85.3} & \bf 89.2 & \bf 69.4 & \underline{55.4} &\underline{64.4}& 49.8 & 55.1 & \underline{65.5}\\

CAF-Net~\cite{CAF-NET} &55.1& 49.3& 57.7 &78.3& 78.6 & 85.1 & 86.2 & 67.4 & 52.0 &\underline{64.4}& 48.3 & 56.2 & 64.9\\
  \midrule

Ours & \bf 64.5 & 	\bf 57.4 & 	54.0 & 	70.9 & 	\bf 80.8 & 	83.6 & 	83.3 & \underline{69.1} & \bf 65.2 & 	\bf 73.0 & 	\bf 50.9 & 	\bf 58.2 & 	\bf 67.6 \\
\toprule
\end{tabular}
}
\label{tab:BP4D}
\end{table*}

\begin{table*}
\centering
\caption{F1 scores, expressed in percentages (\%), are measured for 8 AUs on the DISFA dataset. The best and second best results of each column are indicated with bold font and underlined respectively.}
\resizebox{.7\textwidth}{!}{%
\begin{tabular}{l|cccccccc|c}
\toprule
\multicolumn{1}{l|}{\textbf{Method}} &
  \multicolumn{8}{c|}{\textbf{AU} }&
  \multicolumn{1}{c}{\textbf{Avg.}} \\ 
\multicolumn{1}{l|}{} &
  \multicolumn{1}{c}{1} &
  \multicolumn{1}{c}{2} &
  \multicolumn{1}{c}{4} &
  \multicolumn{1}{c}{6} &
  \multicolumn{1}{c}{9} &
  \multicolumn{1}{c}{12} &
  \multicolumn{1}{c}{25} &
  \multicolumn{1}{c}{26} &
  \multicolumn{1}{|c}{}\\
  \midrule
DRML~\cite{DRML} & 17.3 & 17.7& 37.4 &29.0& 10.7&37.7&38.5&20.1&26.7\\
EAC-Net~\cite{EAC-NET} & 41.5 & 26.4 & 66.4 & 50.7 & \bf 80.5 & \bf 89.3 & 88.9 & 15.6 & 48.5\\
JAA-Net~\cite{JAA} & 43.7 & 46.2 & 56.0 & 41.4 & 44.7 & 69.6 & 88.3 & 58.4 & 56.0\\
SEV-Net~\cite{SEV-net} & \underline{55.3} & 53.1 & 61.5 & 53.6 & 38.2 & 71.6 & \bf 95.7 & 41.5 & 58.8\\
FAUDT~\cite{DBLP:conf/cvpr/JacobS21} & 46.1 & 48.6 & 72.8 & \underline{56.7} & 50.0 & 72.1 & 90.8 & 55.4 & 61.5\\
HMP-PS~\cite{Hybrid} & 38.0 & 45.9 & 65.2 & 50.9 & 50.8 & 76.0 & 93.3 & \underline{67.6} & 61.0\\
ME-AU~\cite{ME-graph} & 54.6 & 47.1 & \underline{72.9} & 54.0 & \underline{55.7} & 76.7 & 91.1 & 53.0 & 63.1\\
CAF-Net~\cite{CAF-NET}  & 45.6 & \underline{55.7} & \bf 80.2 & 51.0 & 54.7 & 79.0 & \underline{95.2} & 65.3 & \underline{65.8}\\
\midrule
Ours & \bf 56.5 & 	\bf 62.6 & 	71.8 & 	\bf {59.2} & 	52.2 & 	\underline{88.0} & 92.4 & 	\bf 73.7 & 	\bf 69.6 \\
\toprule
\end{tabular}
}
\label{tab:DISFA}
\vspace{-0.3cm}
\end{table*}

During fine-tuning process, we employ two widely used FAUs datasets BP4D~\cite{BP4D} and DISFA~\cite{DISFA} to perform FAU detection tasks. BP4D has 328 videos from 41 individuals with around 147,000 frames, and DISFA has 27 videos with 130,815 frames.
We start by identifying the facial region of interest in each dataset and applying cropping and alignment techniques using MTCNN~\cite{MTCNN}. Then we partition the videos into sub-videos comprising 128 frames each, resulting in approximately 1,200 videos as training data. Particularly, AU labels greater than 1 are marked as 1 and the others are labeled as 0, same as previous works~\cite{ME-graph,DRML,EAC-NET}. Following~\cite{ME-graph,DRML,EAC-NET}, we divide the DISFA and BP4D datasets into three folds based on subjects. Two folds are used for training, and the remaining fold is used for testing. The final result is obtained by averaging the test results from the three folds. To improve classification accuracy, data augmentation is used in the DISFA and BP4D datasets to tackle the negative effects of class imbalance, especially the overfitting of certain AU labels. We execute data augmentation by segmenting video clips with rare AU labels and applying random flip and crop operations, resulting in an augmented video dataset, as demonstrated in Fig.~\ref{fig:chart}.


\subsubsection{Evaluation Metrics.}
Performance evaluation for FAU detection commonly relies on both F1 scores and average accuracy (\emph{i.e.} ACC). When dealing with unbalanced samples in a binary classification scenario, the F1 score can more accurately reflect the performance of the algorithm, which is denoted as $F1 = 2\frac{P\cdot R}{P+R}$, where $P$ represents recognition precision and $R$ represents recall rate. In addition to employing the common F1 score as our primary evaluation metric, we have also included an analysis of the detection ACC for each AU label in our supplementary materials.

\subsubsection{Implementation Details.} Both pre-training and fine-tuning networks take in a video consisting of 128 frames resized to $224\times224$ pixels. For pre-training, we use a mask ratio of 0.9 and train the model for 500 epochs with the Adam~\cite{Adam} optimizer. The sampling rate in pre-training is maintained the same as in fine-tuning for data consistency. During fine-tuning, each dataset is trained separately for 20 epochs. For frame-level FAU detection, we adjust the temporal downsampling rate to 4 (selecting one frame every four frames for input into the network). For patch-level FAU detection, we set the mask ratio to 0.5 (randomly masking half of the image patches). Throughout the training and test phases, three varying levels of inputs are utilized, and the resulting output of the network corresponds to the AU label assigned to each frame within the video. The network is trained on 8 NVIDIA 2080Ti GPUs, with an Intel Xeon Gold 6150 @ 2.7GHz CPU.

\subsection{Comparison with State-of-the-art Methods}
\subsubsection{Experiments on BP4D and DISFA Datasets.} For the BP4D dataset, we conducted training and test on 12 AU labels: AU1, AU2, AU4, AU6, AU7, AU10, AU12, AU14, AU15, AU17, AU23, and AU24. The results, presented in Tab. \ref{tab:BP4D}, show that our AU-vMAE model outperformed the state-of-the-art method~\cite{ME-graph} with a 3.2\% increase in F1 score and achieved top performance on 7 AUs.
For the DISFA dataset, we selected 8 AUs~\cite{DRML,EAC-NET,JAA,ME-graph} for training and testing: AU1, AU2, AU4, AU6, AU9, AU12, AU24, and AU25. Table~\ref{tab:DISFA} displays the best F1 scores obtained by our AU-vMAE. The final F1 score shows a promising performance improvement of 5.8\% compared to the state-of-the-art results~\cite{CAF-NET}, with the best performance achieved on 4 AUs.

\begin{figure}[t]
  \centering
  \includegraphics[width=0.95\linewidth]{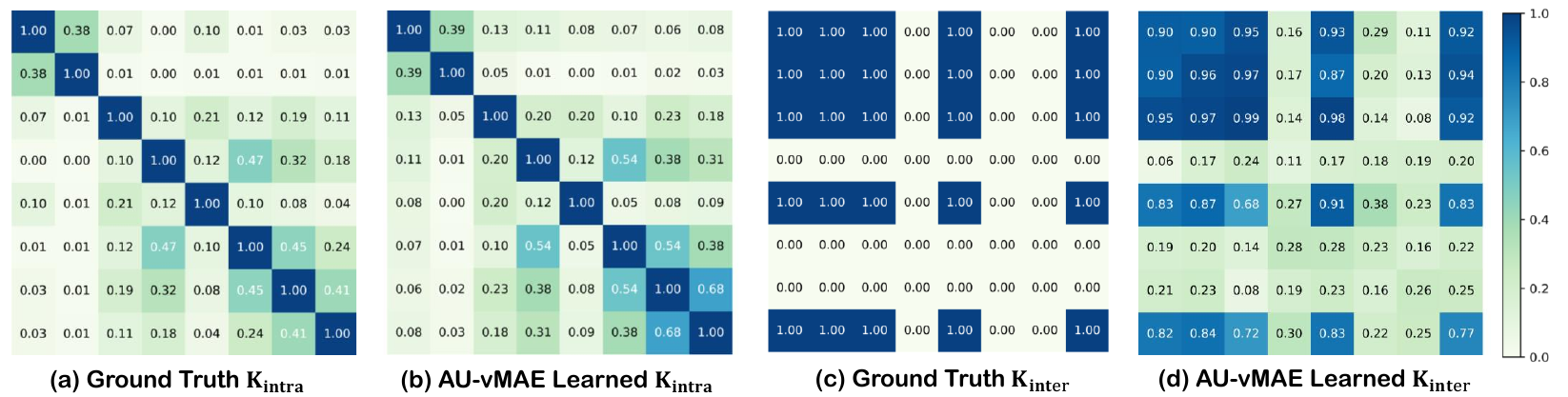}
  \caption{Comparison between ground truth intra-frame (a) and inter-frame knowledge (c), and learned intra-frame (b) and inter-frame knowledge (d).}
  \label{fig:K}
\vspace{-0.2cm}
\end{figure}

In addition to analyzing the F1-score, we conducted the statistical analysis and visual display of the intra-frame and inter-frame knowledge learned from DISFA datasets, as shown in Fig.~\ref{fig:K}. By presenting the state matrix of intra-frame AU pairs and inter-frame AU pairs, we observe high similarity between the learned knowledge and the ground truth knowledge in the test process. 

\subsection{Ablation Study}

\subsubsection{The performance of multi-level FAU detection.} Ablation experiments on three subtasks (video-level, frame-level, and patch-level FAU detection) show varying F1-score values (Tab.~\ref{tab:ablation1}). Taking results on BP4D as examples, video-level detection outperforms frame-level and patch-level detection with a 4.3\% and 9.0\% increase in F1-score, respectively. This is due to missing temporal and spatial information in the latter two subtasks. Despite these limitations, these subtasks demonstrate a commendable level of efficacy in detecting FAUs. These results demonstrate that our AU-vMAE can be utilized to compensate for the lack of spatial and temporal information during detection.
\begin{table}[t]
\centering
\small
\caption{Ablation study: the effects of both intra-frame knowledge and inter-frame knowledge on three subtasks, represented by F1-score (\%) on BP4D and DISFA.}
\resizebox{.9\textwidth}{!}{%
\begin{tabular}{l|ccc|ccc}
\toprule
    \textbf{Dataset} & 
    \multicolumn{3}{c|}{\textbf{BP4D}} &
    \multicolumn{3}{c}{\textbf{DISFA}} 
    \\
  \midrule
\multicolumn{1}{l|}{\textbf{Method}} &
  \multicolumn{1}{c}{video-level } &
  \multicolumn{1}{c}{frame-level } &
  \multicolumn{1}{c|}{patch-level} &
\multicolumn{1}{c}{video-level } &
  \multicolumn{1}{c}{frame-level } &
  \multicolumn{1}{c}{patch-level}
  \\
\midrule
  Baseline & 65.2 & 61.3 & 60.6 & 67.9 & 64.5 & 63.8\\
  Baseline w/ $\mathcal{L}_{intra}$ & 66.8 & 62.5 & 61.7 & 68.5 & 65.9 & 64.7\\
  Baseline w/ $\mathcal{L}_{inter}$ & 65.7 & 64.4 & 61.2 & 68.2 & 66.3 & 64.1\\
  Baseline w/ all & \bf 67.6 & \bf 64.8 & \bf 63.0 & \bf 69.6 & \bf 66.5 & \bf 65.4\\
\toprule
\end{tabular}
}
\vspace{-0.5cm}
\label{tab:ablation1}
\end{table}

\subsubsection{The influence of prior knowledge. }Table~\ref{tab:ablation1} also shows the impact of prior knowledge utilization on AU identification tasks. Four configurations were tested in the ablation experiments. The baseline model used only $\mathcal{L}_{cls}$ as a constraint. $\mathcal{L}_{intra}$ and $\mathcal{L}_{inter}$ were added individually to validate their effectiveness. Finally, both intra-frame and inter-frame knowledge were utilized as constraints. The last row of Tab.~\ref{tab:ablation1} shows the best results among the explored techniques. Incorporating prior statistical knowledge improves detection performance for all three subtasks. The combination of intra-frame and inter-frame knowledge greatly improves AU detection accuracy.

Furthermore, different types of knowledge have varying degrees of influence on different subtasks. Frame-level detection benefits the most from incorporating prior knowledge, with 5.7\% improvements. Video-level and patch-level detection show more modest improvements, with 3.7\% and 4.0\% respectively. 
This aligns with our prior understanding that inter-frame and intra-frame knowledge can compensate for missing timing and space information respectively.

\begin{table}[ht]
\centering
\vspace{-0.2cm}
\caption{Ablation study: the effects of data augmentation shown by F1-score (\%).}
\resizebox{.4\textwidth}{!}{%
\begin{tabular}{l|ccc}
\toprule
\multicolumn{1}{l|}{\textbf{Dataset}} &
  \multicolumn{1}{c}{\textbf{Raw data} } & 
  \multicolumn{1}{c}{\textbf{Augmented data} } \\ 
  \midrule
  DISFA & 69.2 & \bf 69.6 \\
  BP4D & 67.1 & \bf 67.6 \\
\toprule
\end{tabular}
}
\label{tab:ablation2}
\vspace{-0.7cm}
\end{table}

\subsubsection{The influence of data augmentation. }Table~\ref{tab:ablation2} analyzes the impact of data augmentation on model testing outcomes. The augmented data sets show significant performance enhancements of 2.1\% and 1.3\% in F1-score compared to the original datasets. This improvement balances the distribution of AU labels in the augmented data, which prevents overfitting.

\section{Conclusion}
This work proposes AU-vMAE, a videoMAE-based framework for the video-level FAU detection task. It uses a two-stage approach with a pre-trained video feature extractor and a finetuned FAU classification network, which enables video-level, frame-level, and patch-level FAU detection. Additionally, the introduction of intra-frame and inter-frame knowledge priors as a guide in downstream classification tasks and enhances the performance of the model. The experiments conducted demonstrate that pre-training the model effectively reduces the risk of network overfitting and the introduction of prior knowledge can constrain the spatiotemporal relationship of AU pairs, leading to state-of-the-art results on the BP4D and DISFA datasets.

\section*{Acknowledgments}
This work was supported by National Science Foundation of China (U20B2072, 61976137). This work was also partly supported by SJTU Medical Engineering Cross Research Grant YG2021ZD18.

\bibliographystyle{splncs04}
\bibliography{main}
\end{document}


\title{AU-vMAE: Knowledge-Guide Action Units Detection 
via Video Masked Autoencoder \\
\textit{Appendix}}
\titlerunning{AU-vMAE}

%
\author{Paper ID 1211}
\institute{ }
\maketitle              

\section{Additional Experimental Results}

\subsection{ACC Results of AU-vMAE}
Table~\ref{tab:BP4D} and table~\ref{tab:2} exhibit the accuracy outcomes on the DISFA~\cite{DISFA} datasets and the BP4D~\cite{BP4D,BP4D-2} , respectively. Our study's findings reveal a considerable performance improvement when compared to earlier works on FAU detection, as demonstrated across both datasets. Our ACC on the BP4D dataset surpasses previous work by 1.5 percentage points and achieves the best performance on 9 AU labels. Similarly, on the DISFA dataset, we demonstrate a 2.2 percentage point improvement in ACC and achieve the best performance on 5 AU labels.

\begin{table*}[htbp]
\small
\caption{Accuracy, expressed in percentages (\%), is measured for 12 AUs on the BP4D~\cite{BP4D} dataset. Each column's best results are indicated in bold font.}
\begin{tabular}{l|cccccccccccc|c}
\toprule
\multicolumn{1}{l|}{\textbf{Method}} &
  \multicolumn{12}{c|}{\textbf{AU} }&
  \multicolumn{1}{c}{\textbf{Avg.}} \\ 
\multicolumn{1}{l|}{} &
  \multicolumn{1}{c}{1} &
  \multicolumn{1}{c}{2} &
  \multicolumn{1}{c}{4} &
  \multicolumn{1}{c}{6} &
  \multicolumn{1}{c}{7} &
  \multicolumn{1}{c}{10} &
  \multicolumn{1}{c}{12} &
  \multicolumn{1}{c}{14} &
  \multicolumn{1}{c}{15} &
  \multicolumn{1}{c}{17} &
  \multicolumn{1}{c}{23} &
  \multicolumn{1}{c}{24} &
  \multicolumn{1}{|c}{}\\
  \midrule
\textbf{JPML~\cite{JPML}} & 40.7 & 42.1 & 46.2 & 40.0 & 50.0 & 75.2 & 60.5 & 53.6 & 50.1 & 42.5 & 51.9 & 53.2 & 50.5 \\
\textbf{DRML~\cite{DRML}} &55.7 & 54.5 & 58.8 & 56.6 & 61.0 & 53.6 & 60.8 & 57.0 & 56.2 & 50.0 & 53.9 & 53.9 & 56.0 \\
\textbf{EAC-Net~\cite{EAC-NET}} & 68.9 & 73.9 & 78.1 & 78.5 & 69.0 & 77.6 & 84.6 & 60.6 & 78.1 & 70.6 & 81.0 & 82.4 & 75.2 \\
\textbf{ME-AU~\cite{ME-graph}}& 77.7 & 78.1 & 86.5 & \bf 89.2 &  \bf 83.8 & 86.5 & 94.0 &  \bf 73.1 & 84.6 & 78.7 & 80.8 & 86.3 & 83.1 \\
  \midrule
\textbf{Ours} & \bf 82.5 & \bf 83.7 & \bf 90.8 & 86.2 & 77.3 & \bf 88.1 &  \bf 96.6 & 72.5 &  \bf 86.3 &  \bf 80.4 &  \bf 80.5 &  \bf 90.8 &  \bf 84.6\\

\midrule
\end{tabular}

\label{tab:BP4D}
\end{table*}

\begin{table}[t]
    \centering
    \small
    \caption{Accuracy, expressed in percentages (\%), is measured for 8 AUs on the DISFA~\cite{DISFA} dataset. Each column's best results are indicated in bold font.}
    \begin{tabular}{c|cccc|c}
    \toprule
    \textbf{AU} &
    \textbf{JPML} &
    \textbf{DRML}&
    \textbf{EAC-Net}&
    \textbf{ME-AU}&
    \textbf{Ours} \\
    \midrule
    1 & 32.7 & 53.3 & 85.6 & 90.0 & \bf 93.5\\
    2 & 27.8 & 53.2 & 84.9 & 88.5 & \bf 94.4\\
    4 & 37.9 & 60.0 & 79.1 &  94.2 & \bf 95.7 \\
    6 & 13.6 & 54.9 & 69.1 & \bf 92.5 & 92.0\\
    9 & 64.4 & 51.5 & 88.1 & 91.5 & \bf 95.7\\
    12 & 94.2 & 54.6 & 90.0 & 95.9 & \bf 97.4 \\
    25 & 50.4 & 45.6 & 80.5 & \bf 99.1 & 99.3 \\
    26 & 47.1 & 45.3 & 64.8 &  91.2 & \bf 92.9 \\
    \textbf{Avg.} & 46.0 & 52.3 & 80.6 &  92.9 & \bf95.1\\
    \midrule
    \end{tabular}
    \label{tab:2}
\end{table}

Next, we conduct an in-depth analysis of our network's efficacy in comparison to state-of-the-art methods. While the latest methods~\cite{ME-graph,CAF-NET} have proven effective in improving the accuracy of AU detection by incorporating a vast number of learnable parameters (AU relation graphs) into their network architecture, our proposed model enhances accuracy primarily through three distinct aspects. Firstly, we introduce additional data to pre-train the MAE model, enhancing the network's ability to process video features. Secondly, we employ knowledge priors to guide the training of the network, and constrain the state of the AU pairs between frames and within frames, further boosting the network's performance. Thirdly, our proposed model is designed to process videos instead of static images. This enables us to accurately learn state changes in AU labels, while also allowing us to capture temporal correlations and establish consistency within time series data, ultimately leading to an overall improvement in model performance.

\begin{figure}[ht]
  \includegraphics[width=\linewidth]{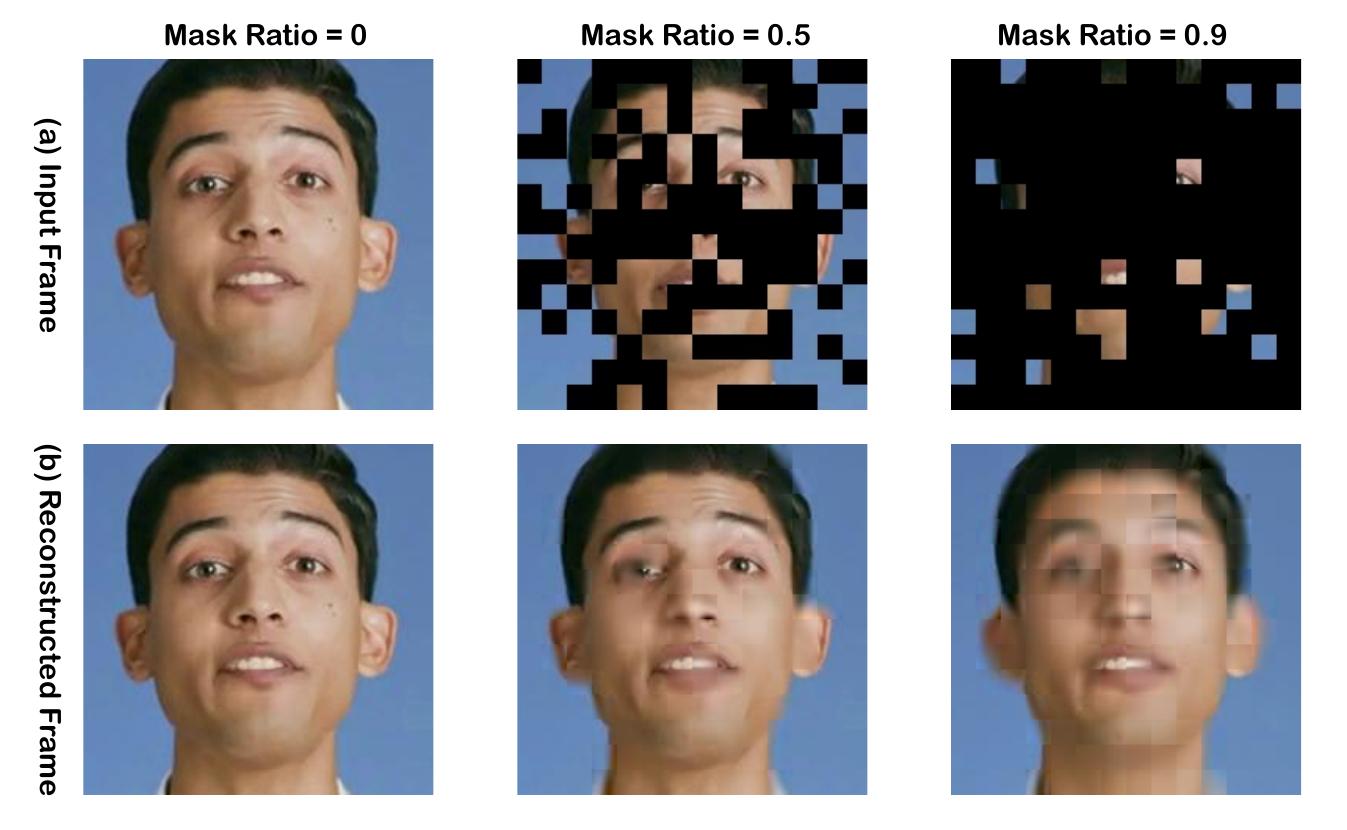}
  \caption{Reconstruction results of the pre-trained model.}
  \label{fig:res}
\end{figure}
\subsection{Reconstruction Results of the Pre-trained Model}
Fig.~\ref{fig:res} illustrates the effectiveness of our pre-trained model in reconstructing missing data under varying mask ratios. During training, we kept the mask ratio fixed at 0.9, while during the test phase, we tested the model's performance on test data with different mask ratios of 0, 0.5, and 0.9. Higher mask ratios have a detrimental impact on the reconstruction quality, as evident from the declining visual fidelity of the final reconstruction results. Despite this, the video frame can still be effectively recovered, indicating our pre-trained model's strong feature extraction capabilities for video data. Moreover, our model demonstrates a superior ability to reconstruct video content even under high mask ratios, which reiterates its efficacy in practical scenarios with missing data. Furthermore, these reconstruction results attest to the feasibility of our subsequent patch-level FAU detection task.

\subsection{Generalizability of AU-vMAE}
To evaluate the generalizability of our model, we conduct cross-dataset tests by training with DISFA and testing on BP4D, and vice versa. We concentrate on the shared AUs—AU1, AU2, AU4, AU6, and AU12—due to differing annotations of these two datasets. The subsequent results, summarized in Tab.~\ref{tab:ablation1}, affirm that the network effectively generalizes across these diverse datasets, despite their intrinsic biases. We appreciate your insightful review and the opportunity to validate our network's robustness in this way.
\begin{table}[t]
\centering
\caption{Cross-dataset Validation.}
\fontsize{9}{12}\selectfont    
\begin{tabular}{llccccc}
\toprule
\multicolumn{1}{c}{\textbf{Training}}&
\multicolumn{1}{c}{\textbf{Test}} & 
\multicolumn{5}{c}{\textbf{FAU}} \\ 
\cmidrule(lr){3-7}
  & & 1 & 2 & 4 & 6 & 12 \\
  \midrule
  BP4D & BP4D \hspace{2.5em}& 64.5 \hspace{2.5em}& 57.4 \hspace{2.5em}& 54.0 \hspace{2.5em}& 70.9 \hspace{2.5em}& 83.3 \\
  BP4D & DISFA \hspace{2.5em}& 61.7 \hspace{2.5em}& 56.1 \hspace{2.5em}& 52.2 \hspace{2.5em}& 68.7 \hspace{2.5em}& 84.4 \\
  DISFA & BP4D \hspace{2.5em}& 55.8 \hspace{2.5em}&  61.8 \hspace{2.5em}& 67.9 \hspace{2.5em}& 59.0 \hspace{2.5em}& 86.1 \\
  DISFA & DISFA \hspace{2.5em}& 56.5 \hspace{2.5em}&  62.6 \hspace{2.5em}& 71.8 \hspace{2.5em}& 59.2 \hspace{2.5em}& 88.0 \\
\toprule
\end{tabular}
\label{tab:ablation1}
\end{table}

\subsection{Computational Efficiency}
 Table~\ref{tab:flops} displays the floating-point operations (FLOPs) per frame of two different input level to facial action unit (AU) detection on frames of resolution $224^2$: video-level detection, which performs detection independently on each frame, and frame-level detection, which incorporates temporal information across frames. We compare these two subtasks to ME-GraphAU~\cite{ME-graph}, with a Transformer backbone. While the Frame-based method processes video data using a complex network (VideoMAE), its computational requirement of 2.81 GFLOPs per frame is lower than ME-GraphAU's 8.57 GFLOPs.
This demonstrates that by exploiting temporal information through our proposed Frame-based approach, we can achieve a reduction in average FLOPs per frame compared to ME-GraphAU, despite using a more powerful backbone network for modeling videos in the FAU detection task.

\begin{table}[ht]
\centering
\caption{Comparison of FLOPs(G) between ME-GraphAU
and our AU-vMAE.}
\begin{tabular}{lccc}
\toprule
\multicolumn{1}{l}{} &
  \multicolumn{1}{c}{ME-GraphAU} & 
  \multicolumn{1}{c}{Video-based} & 
  \multicolumn{1}{c}{Frame-based} \\ 
  \midrule
  GFLOPs & 8.57 & 11.25 & 2.81 \\
\bottomrule
\end{tabular}
\label{tab:flops}
\end{table}

\begin{figure*}[htbp]
  \includegraphics[width=\linewidth]{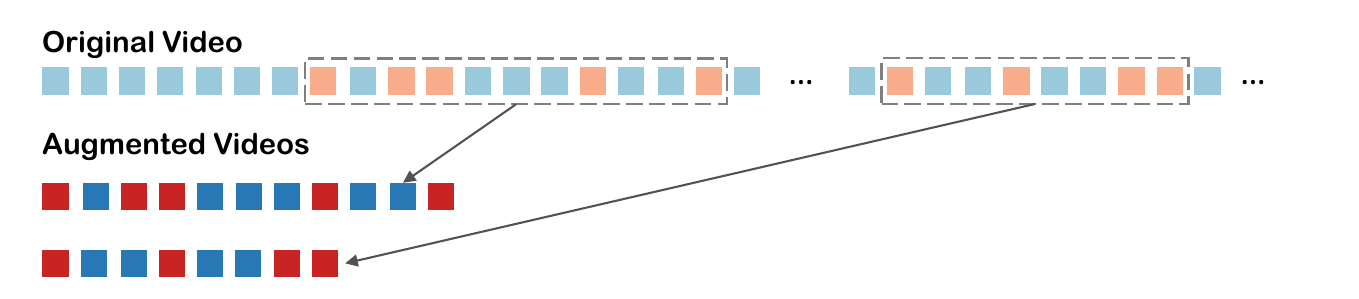}
  \caption{The process of generating augmented video data.}
  \label{fig:aug}
\end{figure*}

\section{Extra Explanation of Data Augment}
In the realm of FAU detection, the identification of FAU is hindered by the extremely unbalanced number of AU labels. To improve the classification accuracy, data augmentation is used in the training data to tackle the negative effects of class imbalance, especially the overfitting of certain AU labels. We execute data augmentation by segmenting video clips with rare AU labels and applying random flip and crop operations, resulting in an augmented video dataset.
The process of generating augmented video data is delineated in Fig.~\ref{fig:aug}. It begins by segmenting AU labels into two categories based on AU frequency: minority AU labels (marked in orange) and majority AU labels (marked in blue). In particular, for the BP4D dataset, the minority AU labels selected consist of AU1, AU2, AU4, AU15, AU23, and AU24. For the DISFA dataset, the selected minority AU labels are AU1, AU2, and AU9. To generate augmented video data, clips are intercepted beginning from the first encountered label and stopped when 15 majority AU labels have been identified consecutively following the minority AU label, as Fig.~\ref{fig:aug} shown.

\bibliographystyle{splncs04}
\bibliography{main}